%% file: m8030.tex
\documentclass{ecai} 
\usepackage{times}  % 设置文档使用的字体为Times Roman
\usepackage{soul}   % 文本高亮、下划线等特殊效果
\usepackage{url}    % 文档中插入URL地址
\usepackage[hidelinks]{hyperref}    % 超链接
\usepackage[utf8]{inputenc} % 设置输入编码UTF-8
\usepackage[small]{caption} % 自定义浮体环境的标题格式
\usepackage{amsmath}    %数学公式
\usepackage{amsthm} % 定义定理、引理、证明等数学结构的样式
\usepackage{booktabs}   % 表格外观
\usepackage{algorithm}  % 编写算法伪代码
\usepackage{algorithmic}    % 与algorithm宏包配合使用
\usepackage[switch]{lineno} % 添加行号

\usepackage{array}
\usepackage[dvipsnames, svgnames, x11names]{xcolor} % 字体颜色
\usepackage{dsfont}   % indicator function正确显示
\usepackage{tabularx}  % 调整表格列宽
\newcolumntype{C}{>{\centering\arraybackslash}X}
\newcolumntype{L}{>{\raggedright\arraybackslash}X}
\newcolumntype{R}{>{\raggedleft\arraybackslash}X}
\usepackage{multirow}   % 创建跨越多行的单元格
\usepackage{wasysym}  % 对号 & 错号
\usepackage{diagbox}    % 斜线表头
\usepackage[toc,page]{appendix} % 附录
\usepackage{graphicx}   % 用于插入图像
\usepackage{subcaption} % 用于管理子图
\usepackage{amssymb}    % 数学公式
\usepackage{amsfonts}   % 数学字体
\usepackage{mathrsfs}   % 数学花体

\newcommand{\BibTeX}{B\kern-.05em{\sc i\kern-.025em b}\kern-.08em\TeX}

\begin{document}

%%%%%%%%%%%%%%%%%%%%%%%%%%%%%%%%%%%%%%%%%%%%%%%%%%%%%%%%%%%%%%%%%%%%%%%%

\begin{frontmatter}

%%% Use this command to specify your submission number.
%%% In doubleblind mode, it will be printed on the first page.

\paperid{8030} 

%%% Use this command to specify the title of your paper.

\title{ReLKD: Inter-Class Relation Learning with Knowledge \\ Distillation for Generalized Category Discovery}

%%% Use this combinations of commands to specify all authors of your 
%%% paper. Use \fnms{} and \snm{} to indicate everyone's first names 
%%% and surname. This will help the publisher with indexing the 
%%% proceedings. Please use a reasonable approximation in case your 
%%% name does not neatly split into "first names" and "surname".
%%% Specifying your ORCID digital identifier is optional. 
%%% Use the \thanks{} command to indicate one or more corresponding 
%%% authors and their email address(es). If so desired, you can specify
%%% author contributions using the \footnote{} command.

\author[A]{\fnms{Fang}~\snm{Zhou}\thanks{Corresponding author. Email: fzhou@dase.ecnu.edu.cn.}}
\author[A]{\fnms{Zhiqiang}~\snm{Chen}}
\author[B]{\fnms{Martin}~\snm{Pavlovski}} 
\author[A]{\fnms{Yizhong}~\snm{Zhang}} 

\address[A]{School of Data Science \& Engineering, East China Normal University, Shanghai, China}
\address[B]{Samsung Electronics America, Mountain View, CA, USA}
% \address[C]{Short Alternate Affiliation of Third Author}

%%% Use this environment to include an abstract of your paper.

\begin{abstract}
Generalized Category Discovery (GCD) faces the challenge of categorizing unlabeled data containing both known and novel classes, given only labels for known classes. Previous studies often treat each class independently, neglecting the inherent inter-class relations. 
Obtaining such inter-class relations directly presents a significant challenge in real-world scenarios.
To address this issue, we propose ReLKD, an end-to-end framework that effectively exploits implicit inter-class relations and leverages this knowledge to enhance the classification of novel classes.
ReLKD comprises three key modules:
a target-grained module for learning discriminative representations, 
a coarse-grained module for capturing hierarchical class relations, and 
a distillation module for transferring knowledge from the coarse-grained module to refine the target-grained module's representation learning. 
Extensive experiments on four datasets demonstrate the effectiveness of ReLKD, particularly in scenarios with limited labeled data. 
The code for ReLKD is available at \href{https://github.com/ZhouF-ECNU/ReLKD}{https://github.com/ZhouF-ECNU/ReLKD}.
\end{abstract}

\end{frontmatter}

%%%%%%%%%%%%%%%%%%%%%%%%%%%%%%%%%%%%%%%%%%%%%%%%%%%%%%%%%%%%%%%%%%%%%%%%

\section{Introduction}\label{sec:introduction}

In recent years, deep learning technology has achieved remarkable progress, driven by the availability of large amounts of labeled data. 
However, acquiring extensive labeled data is often costly and time-consuming.
To effectively leverage unlabeled data, semi-supervised learning (SSL)~\cite{Sohn2020FixMatch,Xie2020Noisy,zhai2019s4l} has emerged as a promising approach. 
Traditional SSL methods typically assume that all classes are present in the labeled data.
However, in real-world scenarios, novel classes often exist within unlabeled data. To address this limitation, emerging research has focused on the problem of Generalized Category Discovery (GCD)~\cite{vaze2022GCD}, where unlabeled data is assumed to contain both known and unknown categories.

The GCD problem requires models to not only recognize known categories\footnote{The terms ``category" and ``class", ``known'' and ``seen'', and ``unknown'' and ``novel'' are used interchangeably.} but also discover new, previously unknown classes within unlabeled data. 
The core challenge lies in the limited availability of supervisory signals for unknown classes.
To address this issue, existing works fine-tune pre-trained models using semi-supervised contrastive learning to obtain discriminative representations.
Subsequently, these methods employ techniques such as semi-supervised clustering~\cite{pu2023DCCL,vaze2022GCD,zhang2023PromptCAL,zhao2023learning} or self-distillation~\cite{ma2024active,peng2024let,wang2024SPTNet,Wen2023SimGCD} to classify both known and unknown  classes.
While these methods have achieved relatively good results, they tend to treat each class independently and overlook the inherent relations between classes. 

\begin{figure}[!t]
    \centering
    % \captionsetup{skip=1pt}
    \includegraphics[width=1\linewidth]{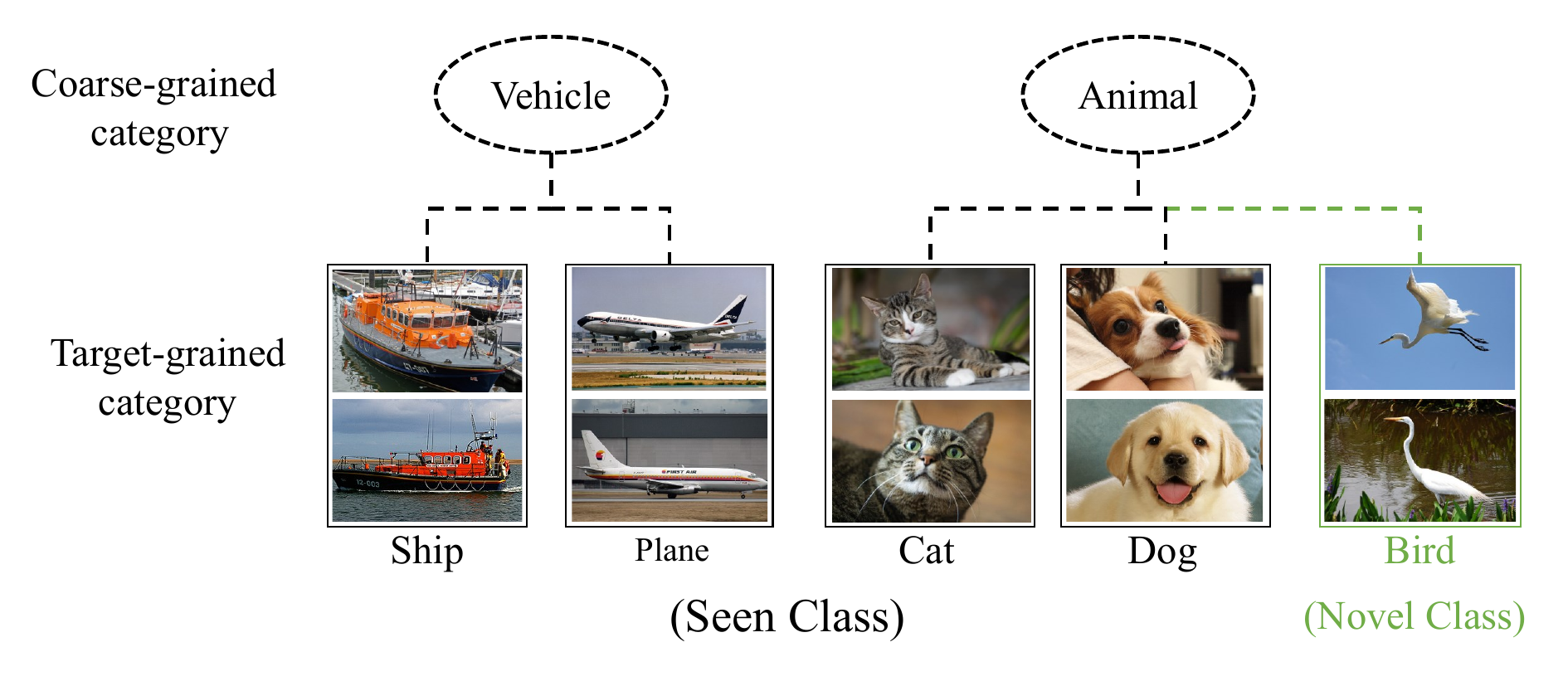}
    \caption{An illustrative example of implicit hierarchical inter-class relations across categories.}
    \label{fig:introduction}
    \vspace{12pt} % 表格下方的空间
\end{figure}

In real-world data, categories often exhibit inherent hierarchical relations and semantic similarities. For instance, as illustrated in Figure~\ref{fig:introduction}, target-level categories like "cat" and "dog" belong to the higher-level coarse-grained category "animal". They share more semantic similarity compared to "ship" and "plane" under the coarse-grained category "vehicle". Consequently, when identifying a novel target-level class like "bird", the target-level categories "cat" and "dog" that belong to the same coarse-grained category "animal" can provide more relevant and valuable information, thereby facilitating a more effective representation learning and further improving classification accuracy. However, in practical applications, the relational and hierarchical information among categories is often difficult to obtain directly. \textit{How to effectively exploit this information and utilize it to enhance the classification performance on target-level categories remains a challenging problem.}

To address the aforementioned challenges, we propose \textbf{ReLKD}, a framework for GCD based on inter-class \textbf{Re}lation \textbf{L}earning with \textbf{K}nowledge \textbf{D}istillation. ReLKD leverages the implicit inter-class relations to improve the learning of target-grained categories by transferring knowledge across different levels of the class hierarchy.
Our framework consists of three key modules: 
(1) \textbf{Target-grained module} learns discriminative representations and prototypes at the target level through semi-supervised contrastive learning and self-distillation, fully utilizing the information from labeled data;
(2) \textbf{Coarse-grained module} aims to learn the implicit hierarchical relations between classes. The challenge lies in the presence of novel classes and the absence of predefined hierarchies. To address this, the coarse-grained module generates pseudo-labels for coarse-grained categories using the labels at the target level, mitigating the lack of direct supervision at the coarse-grained level. Furthermore, it performs contrastive learning between representations and coarse-grained prototypes to capture inter-class similarities and differences. 
(3) \textbf{Distillation module} integrates the information learned at the coarse-grained level into the target level via knowledge distillation, enabling the model to consider the learned hierarchical relations and inter-class relations during classification. In contrast to existing knowledge distillation methods, ReLKD's distillation module transfers knowledge across different hierarchical levels, while performing classification on the more granular (target) level.

In summary, our main contributions are as follows:
\begin{itemize} 
    \item We propose the ReLKD framework to address the GCD problem, which leverages implicit class relations and hierarchical structure information to effectively classify both known and unknown categories. ReLKD not only introduces hierarchical structure information into GCD for the first time, but also effectively exploits hidden inter-class relations to improve novel class discovery.
    \item Recognizing the importance of inter-class relations, we introduce a distillation module to transfer the learned relations towards optimizing the representation learning at the target level.
    %\item ReLKD achieves outstanding performance on four public datasets, \textcolor{red}{provide additional results}.
    \item ReLKD achieves outstanding performance on four datasets, demonstrating a substantial improvement in \textbf{discovering novel classes} while preserving the performance on \textit{seen} classes.
    %effective novel class discovery while preserving the performance on known classes.
\end{itemize}

%%%%%%%%%%%%%%%%%%%%%%%%%%%%%%%%%%%%%%%%%%%%%%%%%%%%%%%%%%%%%%%%%%%%%%%%

% 相关工作
\section{Related Work}\label{sec:relatedWork}

\subsection{Category Discovery}\label{sunsec:catetoryDiscovery}
\textbf{Novel Class Discovery (NCD)} aims to classify instances from an unlabeled set into novel classes. The training data for NCD consists of a labeled set and an unlabeled set. Unlike semi-supervised classification, the unlabeled set in NCD contains entirely distinct class labels that do not overlap with those in the labeled set. 
Earlier methods~\cite{chi2021meta,han2019learning,hsu2018learning,hsu2019multi} were proposed based on the idea of transfer learning, leveraging knowledge learned from known classes to improve the discovery of new classes. 
However, transfer learning-based methods often rely solely on supervised learning with labeled data, overlooking the intrinsic information within unlabeled data. To address this limitation, Zhong et al.~\cite{zhong2021neighborhood} constructed pseudo-positive and hard negative instance pairs by mixing labeled and unlabeled instances to learn richer and more robust feature representations.

\textbf{Generalized Category Discovery (GCD)} extends the NCD problem to a more general setting in which unlabeled data contain both novel and known categories. 
Vaze et al.~\cite{vaze2022GCD} first proposed a two-stage approach combining representation learning and clustering. They fine-tuned a pre-trained vision transformer (ViT) model using semi-supervised contrastive learning, followed by semi-supervised k-means clustering to classify all categories.
Many subsequent works~\cite{pu2023DCCL,zhang2023PromptCAL,zhao2023learning} are built on this framework. 
The aforementioned models are all two-stage approaches based on parameter-free clustering. 
In contrast, end-to-end models based on parametric classification have not performed well in GCD problems and therefore were not widely used initially. 
Wen et al.~\cite{Wen2023SimGCD} first proposed to employ self-distillation for unlabeled data in an end-to-end classification model to produce more balanced pseudo-labels.
Numerous classification models~\cite{cao2024solving,peng2024let,wang2024SPTNet} inspired by~\cite{Wen2023SimGCD} have been developed. 
However, these existing works overlook the relations between classes that can be beneficial for GCD. 

Recent research has also expanded the scope of GCD by addressing more challenging and realistic scenarios, such as domain shifts and class imbalance \cite{bai2023towards,ma2024active,rongali2024CDAD-Net}. Other works~\cite{cendra2024PromptCCD,kim2023proxy,park2025online,wu2023MetaGCD} have integrated GCD with incremental or continual learning, enhancing its relevance to real-world applications. The application of large language models (LLMs) to GCD has also been explored in~\cite{an2023Loop}. These extensions, while significant, are beyond the scope of this paper.

\textbf{Open-world Semi-Supervised Learning (OSSL)} derived from SSL, represents a more realistic learning scenario. 
Essentially, OSSL and GCD share the same problem setting, but OSSL methods~\cite{cao2022ORCA,guo2022robust,ye2023bridging} focus on the differences in the learning process for known and novel categories. 
Although significant progress has been achieved in OSSL, similar to the case with GCD, existing methods often neglect the valuable information embedded within inter-class relations, %This oversight can limit their ability to achieve optimal classification accuracy. 
limitng their classificaiton performance. 
Therefore, one of our main contributions is the effective utilization of implicit inter-class relations to enhance the classification of novel classes.

\subsection{Hierarchical Category Information}\label{subsec:hierCateInformation}
In classification problems, there are hidden hierarchical relations among categories.
Traditional classification methods typically focus on specific levels of a category hierarchy, ignoring the implicit hierarchical relations among categories. Recent works~\cite{an2022fine,rizve2022OpenLDN} have explored the integration of hierarchical information into category discovery. 
Existing methods typically leverage hierarchical information implicitly {during} either the representation learning~\cite{liu2023OpenNCD} or classification~\cite{bai2023towards} stage. 

%As suggested above, 
The concept of hierarchical classification has been explored in other areas, nevertheless its immediate applicability to the GCD problem is limited due to the inherent challenge of novel classes and the absence of predefined hierarchies. To address this challenge, ReLKD incorporates a mechanism that leverages target-grained label information to estimate pseudo-super-class labels. In other words, our work not only introduces hierarchical category information into GCD for the first time, but also effectively exploits hidden inter-class relations, leading to an improved discovery of novel classes. 

\subsection{Knowledge Distillation}\label{subsec:KnowledgeDistillation}
% 知识蒸馏（Knowledge Distillation）旨在通过训练过程中引导学生模型（Student）模仿教师模型（Teacher）所提供的知识，从而提升学生模型的性能。最初，知识蒸馏被应用于模型压缩任务，通过使用参数量更小、结构更轻量的学生模型去拟合教师模型输出的 softmax 预测分布(softmax
% distribution of a teacher model)，以此逼近教师模型在性能上的表现~\cite{hinton2015distilling}。此后，大量研究对蒸馏过程中所使用的知识类型进行了拓展与优化，例如蒸馏中间层的特征表示(features of intermediate layers)~\cite{kim2018paraphrasing,zhou2024rethinking}、注意力图(attention map)~\cite{zagoruyko2017paying}、神经元激活边界(activation boundaries)~\cite{heo2019knowledge}以及样本间的语义关系(instance relation)~\cite{dong2023cluster,park2019relational}等。进一步地，研究者发现知识蒸馏不仅适用于压缩模型，也可用于提升模型性能。例如，相互蒸馏（Mutual Distillation）方法通过让一组未经预训练的学生模型在训练过程中相互协作、互为教师，实现集成式的知识共享，从而提高模型整体性能~\cite{zhang2018deep}。而自蒸馏（Self-Distillation）则采用单一网络在训练过程中同时充当教师与学生，通过自我模仿学习进一步提升模型表达能力与泛化性能~\cite{zhang2019your,dong2023maskclip}。然而，上述知识蒸馏方法通常假设教师和学生针对相同的任务目标和类别集合进行训练，ReLKD使用不同模块对不同层级的分类任务进行学习，并通过内部模块蒸馏的方式将粗粒度层级中的类别相关性蒸馏至目标层级，从而促进在目标层级的分类.

Knowledge distillation aims to improve the performance of a \textit{student} model by guiding it during training to mimic the knowledge provided by a \textit{teacher} model. Initially, knowledge distillation was applied to model compression tasks, where a smaller and more lightweight student model could be trained to approximate the softmax distribution output of a larger teacher model, thereby approaching the teacher's performance~\cite{hinton2015distilling}. Subsequently, numerous studies have expanded and optimized the types of knowledge used in the distillation process, such as features of intermediate layers~\cite{kim2018paraphrasing,zhou2024rethinking}, attention maps~\cite{zagoruyko2017paying}, activation boundaries~\cite{heo2019knowledge}, and semantic relationships between instances~\cite{dong2023cluster,park2019relational}. Furthermore, other works have found that knowledge distillation is not only effective for compressing models but also for enhancing model performance. For example, mutual distillation allows a group of student models (without pre-training) to collaborate and teach each other during training, enabling ensemble-like knowledge sharing that boosts overall performance~\cite{zhang2018deep}. Meanwhile, self-distillation employs a single network that acts as both the teacher and the student during the training process, allowing it to learn by mimicking itself, thereby improving the model's representational capacity and generalization ability~\cite{dong2023maskclip,zhang2019your}. 

The aforementioned knowledge distillation methods typically assume that the teacher and student models are trained for the same task and on the same class sets. ReLKD, in contrast, uses different modules to learn classification tasks at different hierarchical levels and transfers class correlations from a coarse-grained level to the target level through inter-module distillation, thereby promoting classification performance at the target level.

%%%%%%%%%%%%%%%%%%%%%%%%%%%%%%%%%%%%%%%%%%%%%%%%%%%%%%%%%%%%%%%%%%%%%%%%
% 模型框架
\begin{figure*}[!t]
    \centering
    \captionsetup{skip=1pt}
    \includegraphics[width=0.85\linewidth]{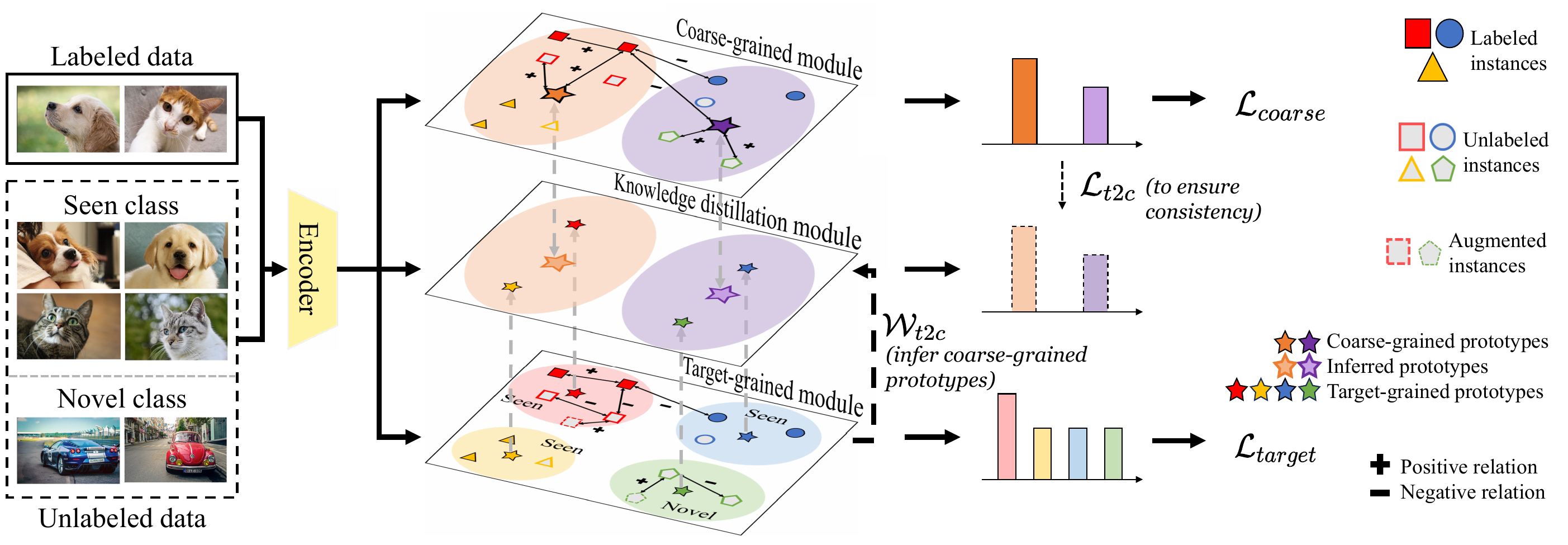}
    \caption{The architecture of ReLKD.}
    \label{fig:modelFramework}
    \vspace{12pt} % 减少表格下方的空间
\end{figure*}

% 方法
\section{Method}\label{sec:method}

\paragraph{Problem Definition.}\label{para:problemDef}
Let $\mathcal{D}$ be a dataset consisting of a labeled set $\mathcal{D}_l=\left\{\left(x_i^l,y_i^l\right)\right\}  \in \mathcal{X} \times \mathcal{Y}_l$, where $\mathcal{Y}_l$ represents the label space of the labeled instances, and an unlabeled set $\mathcal{D}_u=\left\{\left(x_i^u,y_i^u\right)\right\} \in \mathcal{X} \times \mathcal{Y}_u $,  where $\mathcal{Y}_u$ denotes the label space of unlabeled instances.
Note that the unlabeled set $\mathcal{D}_u$ contains instances from both known classes $\mathcal{Y}_l$ and  novel classes $\mathcal{Y}_{novel}$, such that $\mathcal{Y}_u  = \mathcal{Y}_l \cup \mathcal{Y}_{novel}$ and $\mathcal{Y}_l \subset \mathcal{Y}_{u}$. 
The objective of Generalized Category Discovery (GCD) is to learn a model that can assign class labels to instances in  $\mathcal{D}_u$ by leveraging the knowledge in $\mathcal{D}_l$. 

\paragraph{Overall Framework.}\label{para:modelFramework}
The key insight underlying ReLKD is the exploitation of inter-class relations to optimize classification performance. Our framework (as shown in Figure~\ref{fig:modelFramework}) incorporates three key modules: 
(1) a target-grained module learns discriminative representations and prototypes at the target level; 
(2) a coarse-grained module learns the hierarchical relations between classes; and 
(3) a distillation module transfers the knowledge from the coarse-grained module to the target-grained module to optimize representation learning. All modules share a common feature encoder, $f: x \mapsto z \in R^d$, to ensure that the learned representations encompass both coarse-grained and fine-grained information.

\subsection{Target-Grained Module}\label{subsec:TGM}
As a foundation of our framework, we first introduce a target-grained module that (1) employs contrastive learning to learn rich and relevant representations of instances, and subsequently (2) classifies them into learnable prototypes covering both known and novel classes. Following~\cite{vaze2022GCD,Wen2023SimGCD}, we assume that the number of novel classes $|\mathcal{Y}_{novel}|$ is known, and let $K$ represent the number of classes in the dataset $\mathcal{D}$, that is $K = |\mathcal{Y}_u | = |\mathcal{Y}_l | + |\mathcal{Y}_{novel}| $. 

\paragraph{Representation Learning.}\label{para:TGM repreLearning}
To fully exploit the labeled data, we apply supervised contrastive learning~\cite{Khosla2020SCL} on the \textit{labeled} data, and self-supervised contrastive learning~\cite{chen2020simple} on \textit{all} data.
The supervised contrastive loss is 
\begin{equation}
\mathcal{L}_{l}^{rep}=-\frac{1}{\left|B^l\right|} \sum_{i \in B^l} \frac{1}{\left|\mathcal{P}_i\right|} \sum_{p \in \mathcal{P}_i}\log \frac{\exp \left(\boldsymbol{z}_i^{\top} \boldsymbol{z}_p / \tau \right)}{\sum\limits_{j \in B^l} \exp \left(\boldsymbol{z}_i^{\top} \boldsymbol{z}_j / \tau\right)},
%\label{eq:Lrep_l}
\end{equation}
where $\left|B^l\right|$ represents the number of labeled instances in the mini-batch $B$, $\mathcal{P}_i$ is the set of instances having the same label as $x_i$ in $B^l$, and $\tau$ is a  temperature parameter.
For all the instances, the self-supervised contrastive loss is defined as:
\begin{equation}
\mathcal{L}_{all}^{rep}=-\frac{1}{\left|B\right|} \sum_{i \in B} \log \frac{\exp \left(\boldsymbol{z}_i^{\top} \boldsymbol{z}_i^{\prime} / \tau \right)}{\sum\limits_{j \in B} \exp \left(\boldsymbol{z}_i^{\top} \boldsymbol{z}_j / \tau\right)},
%\label{eq:Lrep_all}
\end{equation}
where $\boldsymbol{z}_i^{\prime}$ denotes the representation of a random augmentation of an instance $x_i$.
The representation loss of the target-grained module is $\mathcal{L}_{target}^{rep}=\lambda\mathcal{L}_{l}^{rep} + (1-\lambda)\mathcal{L}_{all}^{rep}$.
% \begin{equation}
% \mathcal{L}_{target}^{rep}=\lambda\mathcal{L}_{l}^{rep} + (1-\lambda)\mathcal{L}_{all}^{rep}.
% \end{equation}

\paragraph{Classification.}\label{para:TGM classification}
A set of learnable prototypes, denoted as $\mathcal{C}=\left\{\boldsymbol{c}_1,\dots,\boldsymbol{c}_{K}\right\}$, is randomly initialized, where each $\boldsymbol{c}_k$ is the learnable representation of class $k$. The probability $\boldsymbol{p}_i^{(k)}$ that an instance $x_i$ belongs to the class $k$ is calculated 
by applying a softmax function to the cosine similarity scores between $z_i$ and the prototypes in $\mathcal{C}$: 
\begin{equation}
\boldsymbol{p}_i^{(k)}=\frac{\exp \left(\operatorname{sim}\left(\boldsymbol{z}_i,\boldsymbol{c}_k\right)\right)}{\sum_{j=1}^{K} \exp \left(\operatorname{sim}\left(\boldsymbol{z}_i,\boldsymbol{c}_j\right)\right)}.
\end{equation}

For \textit{labeled} data, we simply apply the cross-entropy loss between predictions and ground-truth labels, that is, 
\begin{equation}
\mathcal{L}_{l}^{cls} =-\frac{1}{\left|B^l\right|} \sum_{i \in B^l} \boldsymbol{y}_i \log \boldsymbol{p}_i.  
\end{equation}
For \textit{all} the data, we utilize a self-supervised approach, and additionally employ a regularizer to prevent novel classes from being misclassified into a single category:
\begin{align}
\mathcal{L}_{all}^{cls}&=-\frac{1}{\left|B\right|} \sum_{i \in B} \boldsymbol{q}_i^{\prime} \log \boldsymbol{p}_i + \sum_{k=1}^{K} \overline{\boldsymbol{p}}^{(k)} \log \overline{\boldsymbol{p}}^{(k)},
\label{class_all_target}
\end{align}
where $\boldsymbol{q}_i^{\prime}$ is the classification prediction obtained from an augmented version of an instance $x_i$, and $\boldsymbol{p}^{(k)}$ is the average prediction probability for the $k^{th}$ category across all instances in the batch $B$.

The classification loss of the target-grained module is $\mathcal{L}_{target}^{cls}=\lambda\mathcal{L}_{l}^{cls} + (1-\lambda)\mathcal{L}_{all}^{cls}.$
% \begin{equation}
% \mathcal{L}_{target}^{cls}=\lambda\mathcal{L}_{l}^{cls} + (1-\lambda)\mathcal{L}_{all}^{cls}.
% \end{equation}

The overall loss of the target-grained module is as follows:
\begin{equation}
\mathcal{L}_{target}=\mathcal{L}_{target}^{cls} + \mathcal{L}_{target}^{rep}.
\end{equation}

\subsection{Coarse-Grained Module (CGM)}\label{subsec:CGM}
Next, we detail the coarse-grained module, which fulfills two key functions: (1) generating pseudo-super-class labels for labeled data, and (2) employing positive-only contrastive learning to 
enhance the representation learning of labeled instances while simultaneously aligning instance representations with their corresponding super-class prototypes. 
We assume that the number of super-classes, denoted by $K_c$, is known, however, no prior knowledge regarding the super-class label assignments is utilized.

\paragraph{Classification.}\label{para:CGM Classification}

\begin{figure}[!t]
    \centering
    % \captionsetup{skip=1pt}
    \includegraphics[width=1\linewidth]{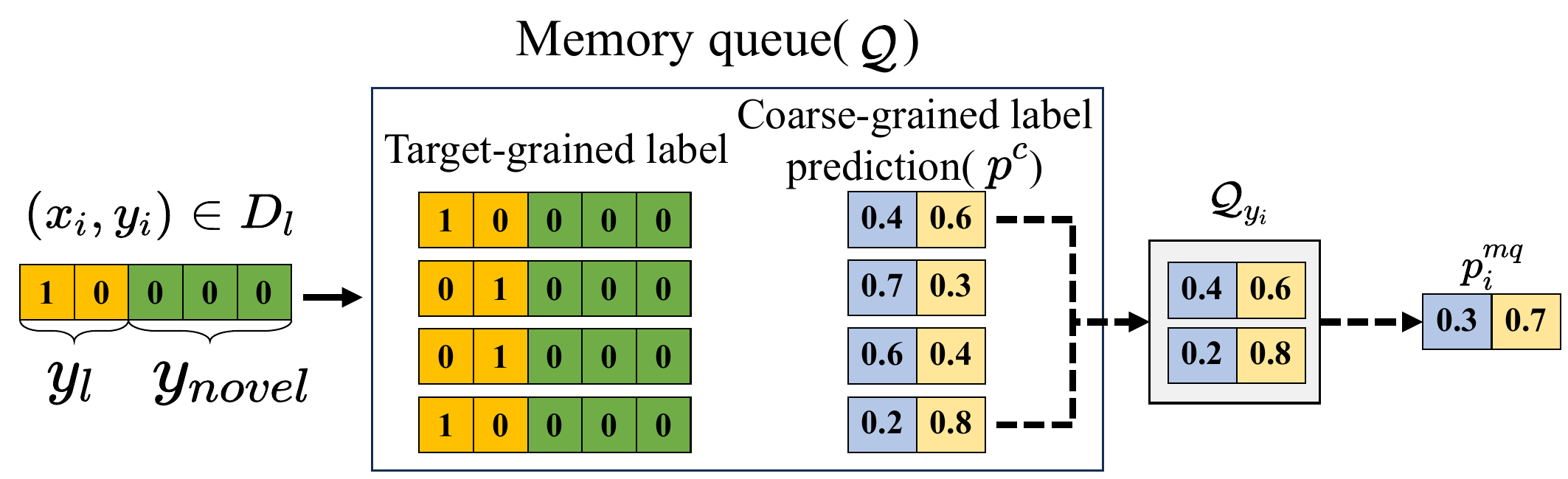}
    \caption{Generation of a pseudo super-class label for a labeled instance.}
    \label{fig:memoryQueue}
    \vspace{15pt} % 表格下方的空间
\end{figure}

Similar to the classification in the target-grained module, we randomly initialize a set of learnable super-class prototypes $\mathcal{C}^c=\left\{\boldsymbol{c}_1^c,\dots,\boldsymbol{c}_{K_c}^c\right\}$, where $\boldsymbol{c}_{k}^c$ is the representation of super-class $k$. The probability $\boldsymbol{p}_i^{c(k)}$ that an instance $x_i$ belongs to the super-class $k$ is computed using the softmax function on the cosine similarity scores between $\boldsymbol{z}_i$ and the super-class prototypes in $\mathcal{C}^c$, that is,
\begin{align}
\boldsymbol{p}_i^{c(k)}=\frac{\exp \left(\operatorname{sim}\left(\boldsymbol{z}_i,\boldsymbol{c}_k^c\right)\right)}{\sum_{j=1}^{K_c} \exp \left(\operatorname{sim}\left(\boldsymbol{z}_i,\boldsymbol{c}_j^c\right)\right)}.
\label{coarseProb}
\end{align}

While we lack explicit super-class labels for individual instances, we can leverage target-grained label information to generate pseudo-super-class labels. Our hypothesis is that instances with the same target-grained labels should also belong to the same super-class. Therefore, for labeled data, we introduce a mechanism to estimate the potential super-class label for each target-grained class $y \in \mathcal{Y}_l$ (see Figure~\ref{fig:memoryQueue}). More specifically,  we maintain a memory queue $\mathcal{Q}$ storing target-grained class labels $y$ and corresponding coarse-grained probabilities $\boldsymbol{p}^c$ (computed using Eq.~\eqref{coarseProb}). Given a labeled instance $x_i$ with class label $y_i$, we select the probabilities $\boldsymbol{p}^c$ associated with the same class label $y_i$ in $\mathcal{Q}$ and store them in  $\mathcal{Q}_{y_i}$.
The pseudo super-class label $\boldsymbol{p}_i^{mq}$ for  instance $x_i$ is then estimated by averaging the  probabilities $\boldsymbol{p}^c$ in the set $\mathcal{Q}_{y_i}$, that is, 
$
    \boldsymbol{p}_i^{mq}=\frac{1}{\left|\mathcal{Q}_{y_i}\right|} \sum_{l \in \mathcal{Q}_{y_i}} \boldsymbol{p}_l^c.
$
Then, the classification loss for the \textit{labeled} data at the coarse-grained level is defined as
\begin{align}
\mathcal{L}_{c\_l}^{cls}&=-\frac{1}{\left|B^l\right|} \sum_{i \in B^l} \boldsymbol{p}_i^{mq} \log \boldsymbol{p}_i^c.
\end{align}
The size of the memory queue $\mathcal{Q}$ is fixed for all experiments. Once a batch of instances has been processed, $\mathcal{Q}$ is updated by storing the class labels $y$ and the corresponding predicted probabilities $\boldsymbol{p}^c$ of the instances in that batch.

For \textit{all} instances, similar to Eq.~\eqref{class_all_target}, we apply a self-supervised approach to calculate the classification loss, that is, 
\begin{equation}
\mathcal{L}_{c\_all}^{cls} =-\frac{1}{\left|B\right|} \sum_{i \in B} \boldsymbol{q}_i^{c\prime} \log \boldsymbol{p}_i^c + \sum_{k=1}^{K_c}  \overline{\boldsymbol{p}}^{c(k)} \log \overline{\boldsymbol{p}}^{c(k)},
\end{equation}
where $\boldsymbol{q}_i^{c\prime}$ represents the predicted super-class label for an augmented version of instance $x_i$.

The total classification loss of the coarse-grained module is $\mathcal{L}_{coarse}^{cls}=\lambda\mathcal{L}_{c\_l}^{cls} + (1-\lambda)\mathcal{L}_{c\_all}^{cls}.$
% \begin{equation}
% \mathcal{L}_{coarse}^{cls}=\lambda\mathcal{L}_{c\_l}^{cls} + (1-\lambda)\mathcal{L}_{c\_all}^{cls}.
% \end{equation}

\paragraph{Representation Learning.}\label{para:CGM repreLearning}
In the absence of true super-class labels, we hypothesize that instances within the same target-grained class are likely to belong to the same super-class. 
We therefore expect representations of instances with identical target-grained labels to be close in the coarse-grained latent space.
To achieve this, we employ a contrastive learning approach without negative pairs, recognizing that instances from different target-grained classes may still share a common super-class:
\begin{align}
\mathcal{L}_{c\_l}^{rep}&=-\frac{1}{\left|B^l\right|} \sum_{i \in B^l} \frac{1}{\left|\mathcal{P}_i\right|} \sum_{p \in \mathcal{P}_i} \operatorname{sim}\left(\boldsymbol{z}_i,\boldsymbol{z}_p\right),
\end{align}
where $\mathcal{P}_i$ represents the set of instances with the same label as $x_i$ in $B^l$.

Then, we employ contrastive learning to align instances' representations with their potential corresponding super-class prototypes. Given an instance $x_i$, let $\hat{y}_i^c=\operatorname{argmax}_k\boldsymbol{p}_i^{c(k)}$ denote its most probable super-class. Considering that the super-class prototype may not be well-learned, the loss is weighted by the probability $\boldsymbol{p}_i^{c(\hat{y}_i^c)}$, i.e.
\begin{align}
\mathcal{L}_{c\_all}^{rep}&=
 -\frac{1}{|B|} \sum_{i \in B} \boldsymbol{p}_i^{c(\hat{y}_i^c)}\log \frac{\exp \left(\boldsymbol{z}_i^{\top}\boldsymbol{c}_{\hat{y}_i^c}^c \right)}{\sum_{k=1}^{K_c} \exp \left(\boldsymbol{z}_i^{\top} \boldsymbol{c}_k^c \right)}.
\end{align}
The representation loss of the coarse-grained module is 
$\mathcal{L}_{coarse}^{rep}=\lambda\mathcal{L}_{c\_l}^{rep} + (1-\lambda)\mathcal{L}_{c\_{all}}^{rep}.$
% \begin{equation}
% \mathcal{L}_{coarse}^{rep}=\lambda\mathcal{L}_{c\_l}^{rep} + (1-\lambda)\mathcal{L}_{c\_{all}}^{rep}.
% \end{equation}

The overall loss of the coarse-grained module is 
\begin{equation}
\mathcal{L}_{coarse}=\mathcal{L}_{coarse}^{cls} + \mathcal{L}_{coarse}^{rep}.
\end{equation}

\subsection{Knowledge Distillation Module (KDM)} \label{subsec:KDM}

Note that the prototypes $\mathcal{C}$ and $\mathcal{C}^c$ are learned independently at the target-grained and coarse-grained levels. Additionally, since $\mathcal{D}$ contains novel classes and the relations between the target-grained labels and super-classes are unknown, the coarse-grained module could not directly improve the classifier's performance at the target-grained level. 
To address this limitation, we introduce a knowledge distillation module, which aims to learn the affinity relations between the prototypes $\mathcal{C}$ and $\mathcal{C}^c$, ensuring that prototypes $\mathcal{C}$ that belong to the same super-class align with their respective $\mathcal{C}^c$. 

%To address this limitation, we introduce a knowledge distillation module to transfer knowledge from the coarse-grained module to the target-grained module by learning the affinity relations between the prototypes $\mathcal{C}$ and $\mathcal{C}^c$.

Specifically, we learn a matrix $\boldsymbol{W}_{t2c}$ that describes the potential relations between target-grained classes and super-classes. Using this matrix, we obtain a set of inferred coarse-grained prototypes $\mathcal{C}^{t2c} =\left\{\boldsymbol{c}_1^{t2c},\dots,\boldsymbol{c}_{K_c}^{t2c}\right\} =\boldsymbol{W}_{t2c} \cdot {\mathcal{C}^T} $.
For an instance $x_i$, the probability of belonging to the inferred super-class $\boldsymbol{c}_{k}^{t2c}$ is computed as: 
\begin{equation}
{\boldsymbol{p}_i^{t2c(k)}}=\frac{\exp \left(\operatorname{sim}\left(\boldsymbol{z}_i,\boldsymbol{c}_k^{t2c}\right)\right)}{\sum_{j=1}^{K_c} \exp \left(\operatorname{sim}\left(\boldsymbol{z}_i,\boldsymbol{c}_j^{t2c}\right)\right)}.
\end{equation}

To ensure consistency between the inferred coarse-grained prototypes $\mathcal{C}^{t2c}$ and learned coarse-grained prototypes $\mathcal{C}^c$, we minimize the cross-entropy loss between coarse-grained predictions $\boldsymbol{p}_i^c $ and inferred predictions $\boldsymbol{p}_i^{t2c}$, defined as:
\begin{align}
\mathcal{L}_{t2c}&=-\frac{1}{\left|B\right|} \sum_{i \in B} \boldsymbol{p}_i^c \log \boldsymbol{p}_i^{t2c} + \sum_{k=1}^{K_c}  {\overline{\boldsymbol{p}}^{t2c(k)}} \log {\overline{\boldsymbol{p}}^{t2c(k)}}.
\label{eq:t2c}
\end{align}
By minimizing the above loss, we establish a reliable relation between target-grained classes and super-classes. 
This strategy ensures effective knowledge transfer from the coarse-grained module to the target-grained module, ultimately improving target-grained classification performance.

\subsection{Overall Loss}\label{subsec:loss}

ReLKD comprises three modules, each with a distinct learning rate to optimize their respective contributions. The target-grained module, leveraging labeled data, initiates the training process to establish a foundation for classification. Subsequently, the coarse-grained module is introduced to refine classification capabilities by exploiting implicit inter-class information. 
Finally, the knowledge distillation module is activated to transfer knowledge from the coarse-grained module to the target-grained module, further boosting the model's classification performance.
The overall loss of ReLKD is defined as 
\begin{align}
\mathcal{L}_{total}=\mathcal{L}_{target} + f_{c}\left(t\right)\mathcal{L}_{coarse} + f_{t2c}\left(t\right)\mathcal{L}_{t2c},
\end{align}
where $f_{c}\left(t\right)$ and $f_{t2c}\left(t\right)$ dynamically adjust the weights of the coarse-grained and distillation modules, respectively, as a function of the training epoch $t$. Specifically, we define early, mid, and late training stages to adjust the coarse-grained and distillation trade-off parameters $f_{c}\left(t\right)$ and $f_{t2c}\left(t\right)$:
\begin{small}
    \begin{equation}
        f_c(t)= 
        \begin{cases}
        0, & 1 \leq t<T_c^{\text {start }} \\ 
        \frac{\lambda_{c}}{2} \cdot \left[1-\operatorname{cos}\left(\frac{t-T_{c}^{\text{start}}}{T_{c}^{\text{end}}-T_{c}^{\text{start}}} \cdot \pi \right)\right], & T_c^{start} \leqslant t<T_c^{\text {end}} \\ 
        \lambda_c, & t \geqslant T_c^{\text {end }}
        \end{cases}
    \end{equation}
    
\end{small}
\begin{small}
    \begin{equation}
        f_{t2c}(t)= 
        \begin{cases}
        0, & 1 \leq t<T_{t2c}^{\text {start }} \\ 
        \frac{\lambda_{t2c}}{2} \cdot \left[1-\operatorname{cos}\left(\frac{t-T_{t2c}^{\text{start}}}{T_{t2c}^{\text{end}}-T_{t2c}^{\text{start}}} \cdot \pi \right)\right], & T_{t2c}^{start} \leqslant t<T_{t2c}^{\text {end}} \\ 
        \lambda_{t2c}, & t \geqslant T_{t2c}^{\text {end}}
        \end{cases}
    \end{equation}
\end{small}
where $T_{c}^{\text{start}}$ and $T_{t2c}^{\text{start}}$ represent the epochs at which the coarse-grained module and distillation module begin contributing to the training process, respectively; $T_{c}^{\text {end}}$ and $T_{t2c}^{\text {end}}$ signify the epochs at which the weights of these two modules stop increasing; and $\lambda_{c} $ and $\lambda_{t2c}$ represent the final weights assigned to the coarse-grained module and distillation module, respectively. 
We employ a cosine function to gradually incorporate the contributions of new modules, initially assigning low weights, progressively increasing them to a peak, and then stabilizing the modules' importance at a certain level.
This strategy ensures a smooth transition and optimal integration of new modules into ReLKD.

%%%%%%%%%%%%%%%%%%%%%%%%%%%%%%%%%%%%%%%%%%%%%%%%%%%%%%%%%%%%%%%%%%%%%%%%

% 实验部分
\input{experiment.tex}

%%%%%%%%%%%%%%%%%%%%%%%%%%%%%%%%%%%%%%%%%%%%%%%%%%%%%%%%%%%%%%%%%%%%%%%%

%%% Use this command to include your bibliography file.

\bibliography{mybibfile}

\end{document}

%% file: experiment.tex
\section{Experiments}\label{sec:experiments}

\subsection{Experimental Setup}\label{subsec:setup}
\paragraph{Datasets.}\label{para:datasets}
To evaluate the effectiveness of ReLKD, four widely used datasets, including CIFAR-100~\cite{krizhevsky2009cifar}, ImageNet~\cite{deng2009imagenet}, Aircraft~\cite{maji2013aircraft} and Scars~\cite{krause2013scars} have been used in our experiments.
\underline{CIFAR-100} includes 100 target-grained classes, which are divided into 20 coarse-grained categories in a balanced manner. Each target-grained class comprises 600 color images of size $32 \times 32$, with 500 of them selected for the training set. 
\underline{ImageNet-100} is a custom subset of ImageNet that we constructed to follow ImageNet's hierarchical label structure~\cite{miller1995wordnet}. Specifically, ImageNet-100 contains 10 target-grained categories selected from each of ImageNet's 10 coarse-grained categories; and for each target-grained category, a training set of 600 images was selected. 

To further validate the performance of ReLKD in \textit{more challenging settings}, we selected the Aircraft and Scars datasets, which focus on finer-grained image categorization tasks with (i) lower variation between target-grained categories and (ii) coarse-grained categories that are not explicitly defined. 
\underline{Aircraft} contains 10,000 images across 100 categories, encompassing various models of airplanes. We allocated two-thirds of all images to the training set, with each class having approximately the same number of images. 
\underline{Scars} includes 16,185 high-resolution images of automobiles across 196 distinct makes, models, and years. %%Being a comprehensive benchmark for fine-grained categorization, 
The dataset has been split by its authors into predefined training and test sets of similar sizes.

Following Vaze's approach~\cite{vaze2022GCD}, we divided the classes from all datasets into \textit{seen} and \textit{novel} categories. For every dataset (except CIFAR-100 which is divided into 80\% \textit{seen} and 20\% \textit{novel} classes), we sampled 50\% of all classes $\mathcal{Y}_u $ as \textit{seen} classes $ \mathcal{Y}_l $. Across the known classes, 50\% of the images are labeled, while the remaining 50\%, along with the images from unknown classes, form the unlabeled set $D_u$. The details of all datasets are shown in Table \ref{tab:dataset}.

\begin{table}[!t]
    % 数据集统计
    \centering
    \scriptsize  
    \captionsetup[table]{skip=1pt} % 减少表格-标题间距
    \vspace{6pt}
    \caption{Datasets' statistics, including the number of \textit{seen} ($\left | \mathcal{Y}_l \right |$) and \textit{novel} ($\left|\mathcal{Y}_u\right|$) categories, the amount of labeled and unlabeled data ($\left|\mathcal{D}_l\right|$ and $\left|\mathcal{D}_u\right|$) as well as the individual sizes of training ($\left|\mathcal{D}_{train}\right|$), validation ($\left|\mathcal{D}_{val}\right|$) and test ($\left|\mathcal{D}_{test}\right|$) sets.}
    \vspace{5pt}
    \setlength{\extrarowheight}{-2pt}
        \begin{tabularx}{0.75\columnwidth}{L R R R R }
        \toprule
        \    & CIFAR-100 & ImageNet-100 & Aircraft & Scars \\ 
        \midrule
        $\left|\mathcal{D}_l\right|$ & 20,000 & 15,000 & 1,666 & 2,036\\
        $\left|\mathcal{D}_u\right|$ & 30,000 & 45,000 & 5,001 & 6,108\\
        $\left|\mathcal{Y}_l\right|$ & 80  & 50 & 50 & 98     \\
        $\left|\mathcal{Y}_u\right|$ & 100 & 100 & 100 & 196   \\ 
        $\left|\mathcal{D}_{train}\right|$ & 50,000 & 60,000 & 6,667 & 8,144 \\
        $\left|\mathcal{D}_{val}\right|$ & 5,000 &  2,500 & 667 & 814 \\
        $\left|\mathcal{D}_{test}\right|$ & 10,000 & 5,000 & 3,333 & 8,041\\ 
        \bottomrule 
    \end{tabularx}
    % \vspace{5pt}
    \label{tab:dataset}
    % \vspace{12pt}  % 减少表格与正文的间距
\end{table}

\paragraph{Baselines.}\label{para:baselines} 
We adapted eight methods for comparison with ReLKD, such as the classical ORCA method~\cite{cao2022ORCA} as well as methods (OpenNCD~\cite{liu2023OpenNCD} and TIDA~\cite{wang2024TIDA}) that leverage multi-level category information in OSSL. We have also considered two-stage clustering-based approaches (GCD~\cite{vaze2022GCD}, DCCL~\cite{pu2023DCCL} and PromptCAL~\cite{zhang2023PromptCAL}) and parametric classification approaches (SimGCD~\cite{Wen2023SimGCD} and LegoGCD~\cite{cao2024solving}). 

\paragraph{Metrics.}\label{para:metrics}
Given that the ground-truth labels of the instances in $D_u$ are not available, we opted to employ Vaze's~\cite{vaze2022GCD} proposed metric for assessing the accuracy of a model's predictions in relation to ground-truth labels:
$
%\begin{equation}\label{Eq 1}
ACC=\max _{p \in \mathcal{P}(y_u)} \frac{1}{\left | D_u \right |} \sum_{i=1}^ {\left | D_u \right |} \mathds{ 1 }\left\{{y_i}=p\left(\hat{y}_i\right)\right\},
%\end{equation}
$
% \mathds{} = \mathbb{}，PDF中能正确渲染
where $\mathcal{P}$ is the set of all permutations computed with the Hungarian algorithm~\cite{kuhn1955hungarian}, and $y_i$ and $\hat{y_i}$ represent the ground-truth label and the clustering prediction from the Hungarian algorithm, respectively. The $ACC$ metric is employed to assess the performance on \textit{Seen}, \textit{Novel}, and \textit{All} classes, separately. 

\begin{table*}[!t]
    % 双栏表
    % CIFAR100 & ImageNet100 & Aircraft & Scars
    % mainResult
    \centering
    \scriptsize % 7pt
    \captionsetup[table]{skip=1pt}
    \caption{$ACC$ of all methods on \textit{All}, \textit{Seen} and \textit{Novel} classes across all datasets. The results of the best-performing methods are \textbf{bolded}, while the second-best results are \underline{underlined}. The results from an additional experiment on CIFAR-100 and ImageNet-100, incorporating ground-truth coarse-grained labels for ReLKD, are included in the bottom row and marked with \textbf{*} if they exceed those of the original ReLKD variant.}
    \vspace{5pt}
    \begin{tabularx}{\textwidth}{@{}L *{12}{>{\centering\arraybackslash}X}@{}}
        \toprule
                                & \multicolumn{3}{c}{CIFAR-100} & \multicolumn{3}{c}{ImageNet-100} & \multicolumn{3}{c}{Aircraft} & \multicolumn{3}{c}{Scars}\\ 
        \midrule
        Methods                 & All & Seen & Novel & All & Seen & Novel & All & Seen & Novel & All & Seen & Novel\\ 
        \midrule
        ORCA                    & 50.9±0.4  & 55.7±0.7 & 31.6±0.9 & 55.2±0.4 & 77.6±0.6 & 32.8±0.9 & 26.0±0.1 & 37.9±0.2 & 24.3±0.1 & 29.1±0.2 & 50.2±0.3 & 16.5±1.0 \\
        OpenNCD                 & 41.2±0.5  & 53.6±0.3 & 33.0±1.1 & 63.2±1.1 & 70.1±1.0 & 56.8±2.1 & 32.2±0.4 & 36.5±1.3 & 27.1±1.6 & 40.2±0.6 & 57.1±1.2 & 25.7±1.3 \\
        TIDA                    & 65.3±0.6  & 73.3±0.5 & 56.6±1.6 & 65.6±0.9 & 69.4±0.6 & 59.3±1.5 & 33.1±0.8 & 37.4±1.1 & 29.7±1.5 & 46.0±0.2 & 55.9±0.7 & 33.5±0.7 \\
        GCD                     & 72.1±0.3  & 76.3±0.6 & 53.4±2.8 & 68.1±1.2 & 76.1±2.3 & 60.0±3.4 & 40.2±0.7 & 50.2±1.7 & 30.3±2.2 & 42.9±0.6 & 64.7±0.3 & 21.9±1.3 \\
        DCCL                    & 74.2±1.2  & 80.4±1.6 & 49.4±3.8 & 66.8±0.9 & 72.2±1.6 & 61.3±1.9 & 36.8±0.7 & 41.1±1.5 & 32.5±1.3 & 33.3±0.2 & 43.3±0.7 & 23.7±0.7 \\
        PromptCAL               & 81.6±0.2  & \textbf{85.8±0.4} & 70.1±2.4 & 73.3±1.3 & 78.9±1.4 & 67.6±2.1 & 53.6±0.6 & 57.6±0.9 & 49.6±1.3 & 53.8±0.6 & 70.3±1.4 & 37.8±1.0 \\
        SimGCD                  & 80.2±0.6  & 81.0±0.4 & 77.6±2.1 & 72.3±0.5 & 78.4±0.2 & \underline{67.9±1.3} & \underline{54.3±1.4} & \underline{59.3±0.4} & \underline{51.6±0.9} & 53.5±1.5 & 71.4±1.1 & \underline{44.7±1.8} \\ 
        LegoGCD                  & \underline{82.1±0.2}  & \underline{82.7±0.3} & \underline{80.6±1.1} & \underline{73.5±0.2} & \underline{84.6±0.9} & 62.4±0.5 & 53.8±0.4 & 58.0±0.7 & 50.1±0.2 & \textbf{59.3±0.2} & \textbf{74.8±0.6} & 44.3±0.2 \\  
        \midrule
        ReLKD                    & \textbf{82.6±0.3}  & 82.1±0.2 & \textbf{84.5±1.0} & \textbf{78.4±0.5} & \textbf{84.8±0.4} & \textbf{72.0±0.9} & \textbf{56.7±0.1} & \textbf{60.5±1.0} & \textbf{52.9±1.0} & \underline{58.7±0.5} & \underline{72.4±1.0} & \textbf{45.6±0.8} \\ 
        ReLKD$_{+GT_c}$            & \textbf{84.6±0.1*} & 83.6±0.2 & \textbf{86.4±0.4*} & \textbf{80.3±0.7*} & \textbf{85.3±0.5*} & \textbf{75.2±1.2*} & - & - & - & - & - & - \\
        \bottomrule
    \end{tabularx}
    \label{tab:mainResults}
    \vspace{10pt} % 表格下方的空间
\end{table*}

\paragraph{Implementation details.} \label{para:details}
In the experiments, we utilized the pre-trained ViT-B/16 as the backbone of ReLKD's encoder and conducted fine-tuning exclusively of the final block of the vision transformer. The optimal hyperparameter values for each component of ReLKD were obtained by grid search. In accordance with the previous study by Vaze and Wen et al.~\cite{vaze2022GCD,Wen2023SimGCD}, we used the number of target-grained categories $\left |\mathcal{Y}_u \right|$ as a priori information, i.e. as the number of prototypes at the target level $K = \left |\mathcal{Y}_u \right|$. As for the coarse-grained level in CIFAR-100 and ImageNet-100, we directly utilized the number of coarse-grained categories as a priori information to determine the number of coarse-grained prototypes $K_c$. 
For Aircraft and Scars, the two datasets without explicit coarse-grained category labels, we experimented with different numbers of  $K_c$ using the validation set; and identified that optimal results are obtained with 20 and 30 coarse-grained categories, respectively. 
ReLKD was trained for 200 epochs and optimized with the SGD optimizer. All baselines were run using their original implementations as described in their respective papers, and their hyperparameter values were selected via grid search. All experiments were conducted on a server machine equipped with an Intel(R) Xeon(R) Gold 6240R CPU, a Tesla V100-SXM2-32GB GPU, and 256 GB of RAM.

\subsection{Effectiveness of ReLKD}\label{subsec:mainResult}
Table \ref{tab:mainResults} presents the $ACC$ of ReLKD and the baselines across the four datasets. Overall, ReLKD achieves superior performance in classifying \textit{All}, \textit{Seen} and \textit{Novel} classes in the majority of cases, with suboptimal results observed in only three comparisons. These results demonstrate the superior effectiveness of ReLKD. % 加入新baseline后有3处没有达到最优
In particular, ReLKD improves the $ACC$ for \textit{All} classes by 0.5\% and 4.9\% on CIFAR-100 and ImageNet-100, respectively. Moreover, it achieves considerable respective improvements of 3.9\% and 4.1\% in the $ACC$ for \textit{Novel} classes. Notwithstanding these improvements, ReLKD still lags behind PromptCAL and LegoGCD in terms of its performance for \textit{Seen} classes on CIFAR-100. 
This is due to some coarse-grained categories in CIFAR-100 having minimal differences, such as "vehicles 1" and "vehicles 2", thus limiting the effectiveness of the coarse-grained module in enhancing target-level classification. 
Apart from this minor downside, ReLKD outperforms these two models on all other datasets. %%%demonstrating ReLKD's superior class discovery performance while maintaining the ability to better categorize known classes.

Furthermore, the results on the more challenging datasets (Aircraft and Scars) demonstrate that ReLKD also exhibits improved performance. With the exception of the accuracy on the \textit{All} classes and \textit{Seen} classes in  Scars, which is slightly lower than that of LegoGCD, the remaining metrics reach the current state-of-the-art results. It is worth noting that, since the number of coarse-grained classes in Aircraft and Scars are not provided, ReLKD uses an estimated number of coarse-grained classes and still achieves substantial improvements.

Due to differences in dataset characteristics such as class granularity and inter-class structure complexity, it is reasonable that the performance gains vary across datasets with some coarse-grained classes not being explicitly defined. Note that such variability is not unique to ReLKD and impacts the baseline methods as well, whose performances also vary across datasets due to the same factors. Nevertheless, despite such dataset-specific variations, ReLKD consistently outperforms the baselines.

Finally, recognizing that ReLKD's Coarse-Grained Module (CGM) relies solely on a priori knowledge of the predefined number of coarse-grained categories, we introduce an additional experiment. 
This experiment, conducted on CIFAR-100 and ImageNet-100, provides an upper bound on ReLKD's performance when accurate coarse-grained labels ($GT_c$) are available. 
The results presented in the last row of Table \ref{tab:mainResults} demonstrate a notable performance improvement when $GT_c$ is incorporated,
suggesting that learning with coarse-grained information plays a pivotal role in GCD.

\begin{figure*}[t!]
    % 组合图
    \centering
    \captionsetup[table]{skip=1pt}
    \begin{subfigure}{0.195\textwidth}
        \includegraphics[width=\linewidth]{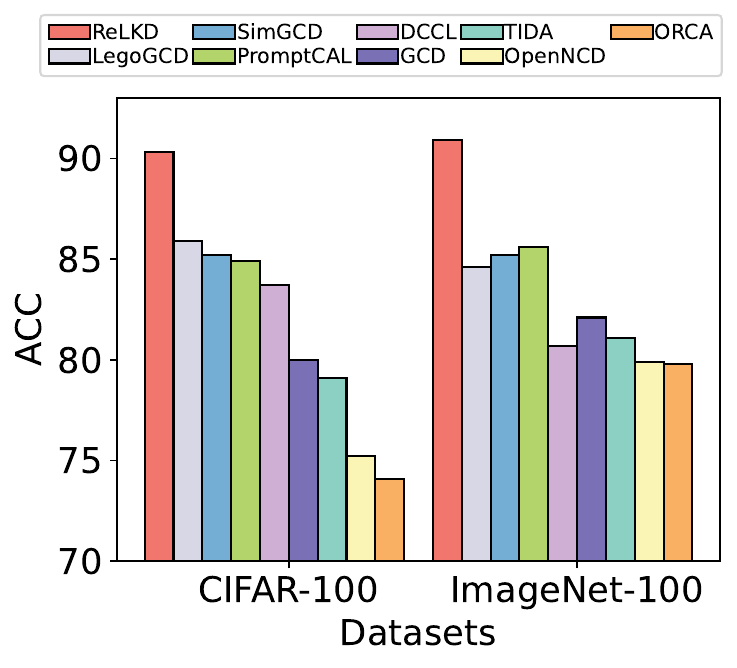}
        \caption{\scriptsize $ACC$ of all methods at the coarse-grained level.}
        \label{fig:coarseGrainedEffectiveness}
    \end{subfigure}
    \hfill
    \begin{subfigure}{0.19\textwidth}
        \includegraphics[width=\linewidth]{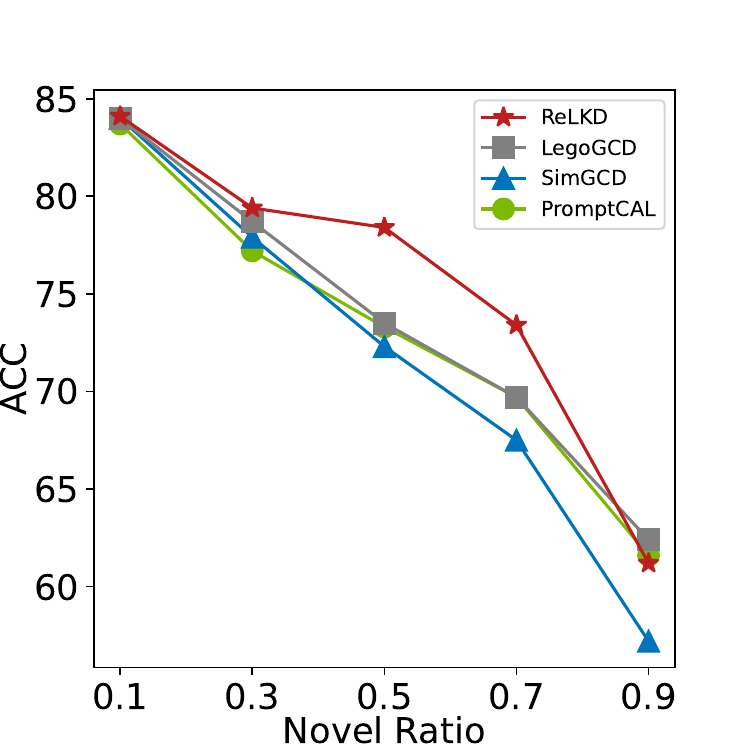}
        \caption{\scriptsize $ACC$ on \textit{All} Classes}
        \label{fig:NR all}
    \end{subfigure}
    \hfill
    \begin{subfigure}{0.19\textwidth}
        \includegraphics[width=\linewidth]{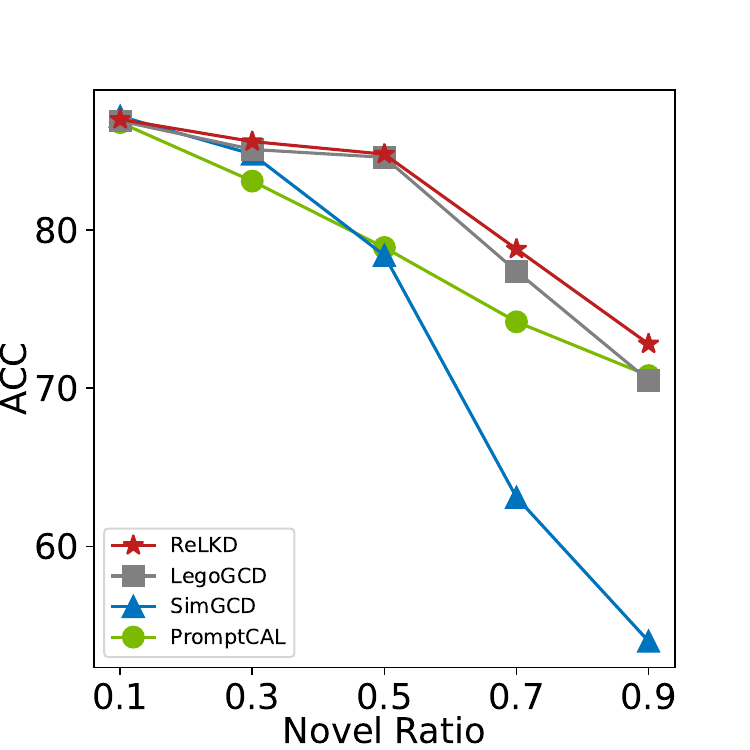}
        \caption{\scriptsize $ACC$ on \textit{Seen} Classes}
        \label{fig:NR seen}
    \end{subfigure}
    \hfill
    \begin{subfigure}{0.19\textwidth}
        \includegraphics[width=\linewidth]{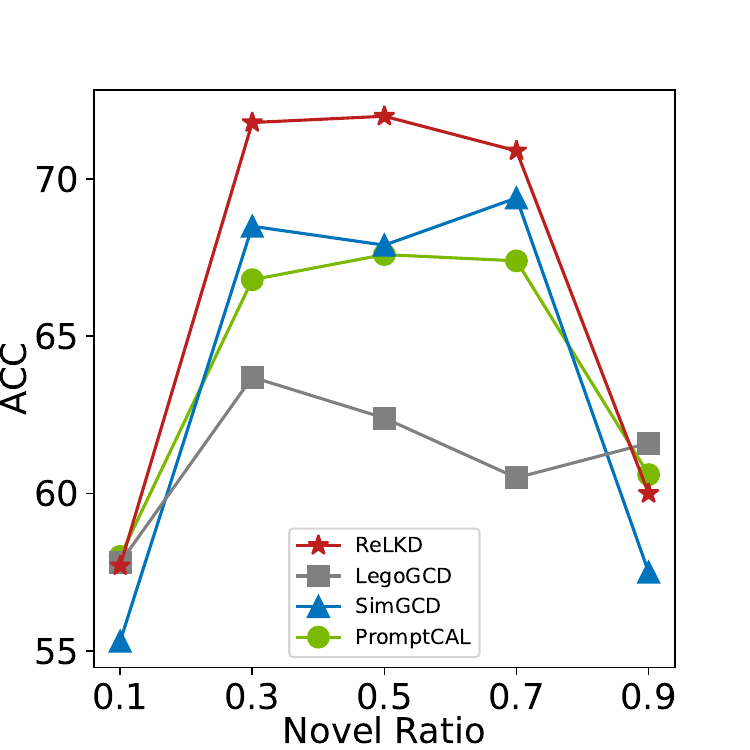}
        \caption{\scriptsize $ACC$ on \textit{Novel} Classes}
        \label{fig:NR novel}
    \end{subfigure}
    \hfill
    \begin{subfigure}{0.2\textwidth}
        \includegraphics[width=\linewidth]{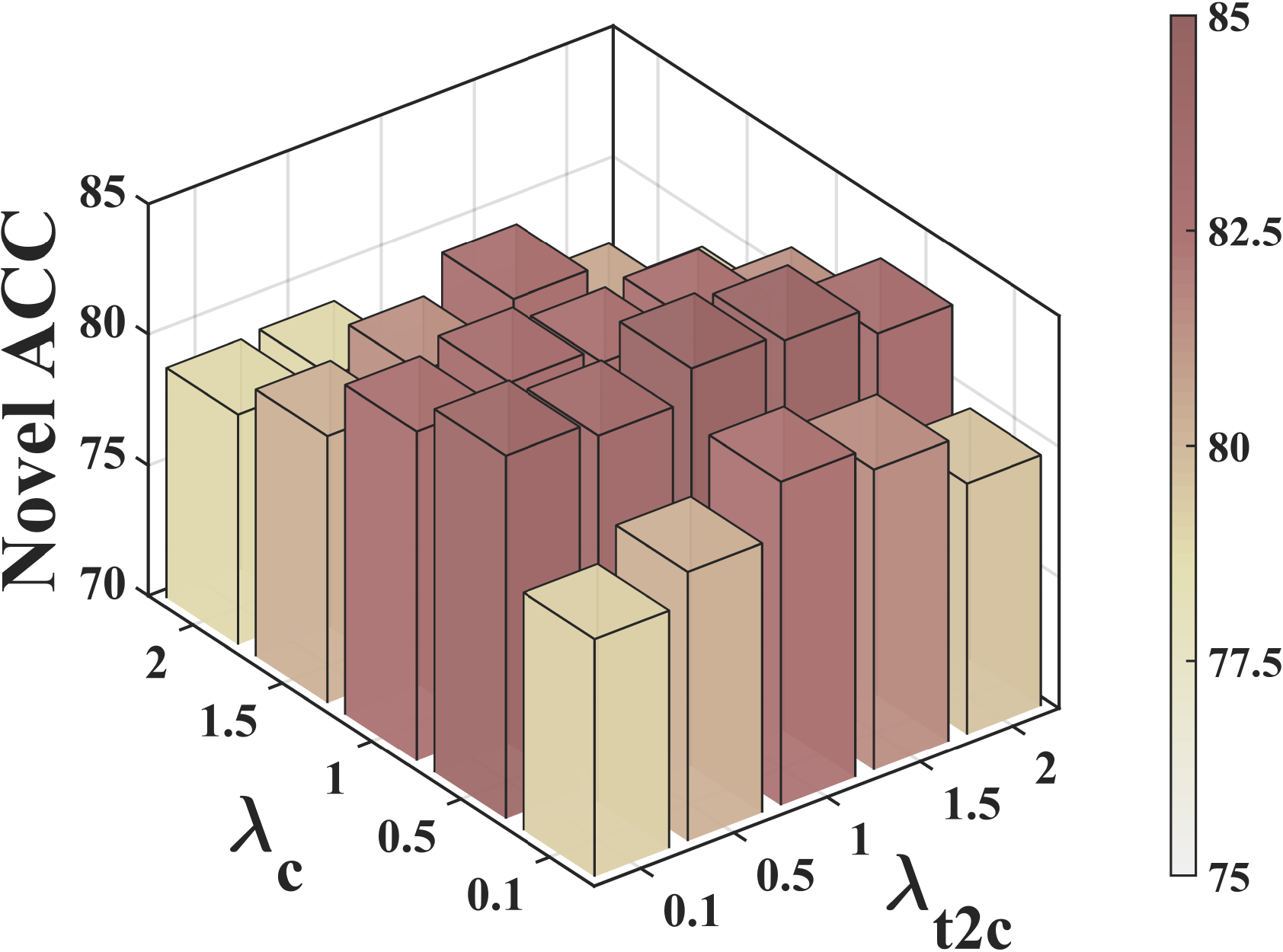}
        \caption{\scriptsize $ACC$ on CIFAR-100 for different loss weights.}
        \label{fig:lossWeights}
    \end{subfigure}
    \vspace{10pt} % 减少表格下方的空间
    \caption{
    (a) Coarse-grained ACC. (b-d) Impact of novel ratio on \textit{All}, \textit{Seen}, and \textit{Novel} classes. (e) Impact of loss weights on \textit{Novel} ACC.}
    \label{fig:furtherAnalysis}
    \vspace{10pt} % 减少表格下方的空间
\end{figure*}

\subsection{Evaluation of Coarse-Grained Performance}\label{subsec:cGrainedEff}
\paragraph{Effectiveness.}\label{para:cGEff} 
To evaluate the performance of ReLKD at the coarse-grained level, we conducted experiments by comparing the coarse-grained classification of ReLKD and all baselines on CIFAR-100 and ImageNet-100, both of which provide coarse-grained labels. 
Specifically, we mapped all methods' target-grained predictions to their corresponding coarse-grained categories, and then computed their $ACC$s. 
A prediction was deemed correct at the coarse-grained level if the predicted target-grained category belonged to the same coarse-grained category as the ground-truth target-grained category.
For example, for an instance with the ground-truth label ``cat'' at the target-grained level, if the model classifies it as "cat", "dog", or any other category within the ``animal" coarse-grained category, then the classification is considered correct \textit{at the coarse-grained level}.

Figure \ref{fig:coarseGrainedEffectiveness} shows that ReLKD obtains relatively large improvements compared with other baselines, i.e. 4.4\% and 5.3\% on CIFAR-100 and ImageNet-100, respectively. Meanwhile, the results in Table \ref{tab:mainResults} and Figure \ref{fig:coarseGrainedEffectiveness} are fairly consistent, which indicates that improving the classification ability at the coarse-grained level can help to indirectly improve the performance at the target-grained level.

\paragraph{Sensitivity to coarse-grained prototype number.}\label{para:numOfKc} 
To evaluate the sensitivity of ReLKD to the number of predefined coarse-grained categories, we varied $K_c$ and calculated the corresponding $ACC$ on CIFAR-100 and ImageNet-100. Table \ref{tab:numOfKc} shows that ReLKD is able to achieve superior results when $K_c$ is close to the true number of coarse-grained classes. In addition, even when $K_c$ differs significantly from the actual number of coarse-grained categories, ReLKD still outperforms SimGCD which does not utilize coarse-grained information. These results indicate that (1) ReLKD is not sensitive to $K_c$, and (2) capturing correlation between categories at the target level indeed enhances the classification performance.

\begin{table}[!t]
    % CIFAR100 和 ImageNet100 不同粗粒度类别数量对比结果
    \scriptsize
    \centering
    \captionsetup[table]{skip=1pt}
    \caption{$ACC$ computed for different $K_c$ values on CIFAR-100 and ImageNet-100. The value marked with $*$ represents the true coarse-grained quantity in that dataset.}
    \vspace{5pt}
    \begin{tabularx}{\columnwidth}{L@{}CCC L@{}CCC}
        \toprule
        \multirow{2}{*}{$K_c$} & \multicolumn{3}{c}{CIFAR-100} & \multirow{2}{*}{$K_c$} & \multicolumn{3}{c}{ImageNet-100} \\ 
            & All & Seen & Novel &      & All & Seen & Novel \\ 
        \midrule
        10 & 80.8±0.5  & 81.5±0.2 & 78.4±1.2 & 5 & 76.1±0.5 & 81.9±0.2 & 70.2±0.7 \\
        15 & 81.2±0.3  & 81.4±0.4 & 80.3±0.9 & $10^*$ & \textbf{78.4±0.5} & \underline{84.8±0.4} & \textbf{72.0±0.9} \\
        $20^*$ & \textbf{82.6±0.3} & \textbf{82.1±0.2} & \textbf{84.5±1.0} & 15 & \underline{78.4±0.9} & \textbf{85.1±0.8} & \underline{71.8±1.2} \\
        25 & \underline{82.1±0.4} & 81.5±0.1 & \underline{83.9±0.7} & 20 & 77.1±0.4  & 83.6±0.3 & 70.8±0.6 \\
        30 & 81.7±0.2  & \underline{81.6±0.2} & 82.0±1.0 & 25 & 77.2±0.6  & 84.1±0.7 & 69.2±0.5 \\
        40 & 81.3±0.6  & 81.5±0.3 & 80.3±0.6 & 30 & 76.1±0.3 & 81.6±0.3 & 69.4±0.7 \\ 
        \bottomrule
    \end{tabularx}
    \vspace{5pt} % 表格下方的空间
    \label{tab:numOfKc}
\end{table}

\subsection{Ablation Study: Component Contributions}\label{subsec:ablationStudy}
To ascertain the veracity of the individual modules in ReLKD and to determine their respective contributions, we conducted an ablation study on CIFAR-100. The study focuses on analyzing the proposed CGM and KDM,  and the results are shown in Table \ref{tab:ablation}. 

Compared to SimGCD, which uses only the target-grained module, adding the coarse-grained module (CGM) (Section~\ref{subsec:CGM}), improved performance by 1.6\%, 0.9\%, and 4.0\% on \textit{all}, \textit{seen}, and \textit{novel} classes, respectively. Further incorporating the knowledge distillation module (KDM) (Section~\ref{subsec:KDM}) yielded additional improvements of 0.8\%, 0.2\%, and 2.9\% on \textit{all}, \textit{seen}, and \textit{novel} classes, respectively. These outcomes clearly demonstrate the substantial impact of each module on ReLKD's overall performance. 

The introduction of the CGM allows for the indirect influence of coarse-grained information on ReLKD’s representation learning, and the introduction of KDM enables the knowledge transferred from CGM to guide the learning of prototypes at the target level, thereby enhancing the ability of ReLKD for novel class discovery. Furthermore, we also found that both CGM and KDM result in a notable enhancement in the model's performance on \textit{Novel} classes. This observation suggests that ReLKD has effectively exploited the correlation between target-grained level classes through coarse-grained information, facilitating the transfer of knowledge from \textit{Seen} classes to \textit{Novel} classes in a more effective manner.

\begin{table}[!t]
    % 消融实验
    \centering
    \scriptsize
    \setlength{\extrarowheight}{-2pt} % 减少行间距
    \captionsetup[table]{skip=1pt}
    \vspace{5pt}
    \caption{$ACC$ on CIFAR-100 for ablation study.}
    \vspace{5pt}
    \begin{tabularx}{1\columnwidth}{@{}C CCC CCC@{}}
        \toprule
        \multirow{2}{*}{Variant} & \multirow{2}{*}{CGM}  & \multirow{2}{*}{KDM} & \multicolumn{3}{c}{CIFAR-100} \\
                                 &                      &                      & All       & Seen      & Novel      \\ 
        \midrule
        SimGCD                 & \XBox               & \XBox               & 80.2±0.6  & 81.0±0.4 & 77.6±2.1 \\
        ReLKD$_{\text{-KDM}}$                & \checked            & \XBox               & \underline{81.8±0.4}  & \underline{81.9±0.4} & \underline{81.6±0.9} \\
        ReLKD               & \checked            & \checked            & \textbf{82.6±0.3}  & \textbf{82.1±0.2} &\textbf{84.5±1.0} \\
        \bottomrule
    \end{tabularx}
    \vspace{5pt} % 减少表格下方的空间
    \label{tab:ablation}
\end{table}

\begin{table}[!t]
    % 粗粒度Warm-up
    \scriptsize
    \centering
    \captionsetup[table]{skip=1pt}
    \caption{Warmup experiments for CGM.}
    \vspace{5pt}
    \begin{tabularx}{1 \columnwidth}{@{} CC CCC @{}}
        \toprule
        $T_{c}^{start}$ & $T_{c}^{end}$ & All & Seen & Novel \\ 
        \midrule
        0 & 30 & 74.6  & 79.6 & 54.4 \\
        30 & 45 & \underline{82.4}  & \underline{83.1} & 79.6 \\
        30 & 60 & \textbf{82.6}  & 82.1 & \textbf{84.5} \\
        30 & 90 & 81.3  & 81.4 & \underline{81.3} \\
        60 & 90 & 81.8  & \textbf{83.4} & 76.2 \\ 
        \bottomrule
    \end{tabularx}
    % \vspace{5pt} % 减少表格下方的空间
    \label{tab:CGMwarmUp}
\end{table}

\subsection{Effect of Novel Class Ratio}\label{subsec:novelRatio}

We evaluated the performance of ReLKD on ImageNet-100 under different novel class ratios, ranging from 0.1 to 0.9. The classification accuracies for \textit{All}, \textit{Seen}, and \textit{Novel} classes are shown in Figure~\ref{fig:furtherAnalysis}.
The results demonstrate that ReLKD consistently outperforms LegoGCD, SimGCD and PromptCAL (the three leading baselines in Table \ref{tab:mainResults}) across all novel class ratios, showcasing its robustness in diverse scenarios.
Specifically, ReLKD exhibits a significant advantage when the novel class ratio is 0.3, 0.5 and 0.7.  
When the novel ratio is 0.1, the number of instances from unknown classes is limited, making performance improvements challenging. Conversely, when the novel ratio is 0.9, the number of instances from known classes is significantly reduced, leading to insufficient learning from known classes, thereby limiting the generalization ability to unknown classes.
% 由于三张Novel Ratio图都合并到图组中，具体是(b),(c),(d)三张子图。引用可能需要做出改变。

\subsection{Parameter Impact Analysis}\label{subsec:parameterImpact}
This section analyzes the impact of specific parameters utilized in our method. We conducted two sets of experiments to explore 1) the loss weights of CGM and KDM, and 2) the start and duration of the loss weight growth during training. 

\paragraph{Loss weights.}\label{para:lossWeights} In order to optimize the performance of the model, it is first necessary to determine the loss weights (denoted as $\lambda_c$ and $\lambda_{t2c}$) of CGM and KDM. The optimal $\lambda_c$ and $\lambda_{t2c}$ were identified based on the $ACC$ for the \textit{Novel} classes on the CIFAR-100 validation set. The obtained results are shown in Figure \ref{fig:lossWeights}. When $\lambda_c$ and $\lambda_{t2c}$ are small, the model has limited improvement and cannot effectively use the learned coarse-grained information, while when $\lambda_c$ and $\lambda_{t2c}$ are large, there is a decreasing trend in $ACC$. This may be due to the fact that the goal of the experiment is to classify categories at the target-grained level, and very large weights for CGM and KDM may have a negative effect as the classification ability deviates from the target-grained level.

\paragraph{Warm-up.}\label{para:warmUp} After determining $\lambda_c$ and $\lambda_{t2c}$, we also need to know when additional modules should be introduced into ReLKD. We set different start/end time points for CGM (denoted as $T_{c}^{start}$ and $T_{c}^{end}$) on the CIFAR-100 validation set and calculate the corresponding $ACC$s. Table \ref{tab:CGMwarmUp} shows that if CGM is introduced too early, the model will suffer insufficient learning at the target level, which leads to a serious performance degradation. However, the time of introducing KDM has little effect on the performance of ReLKD\footnote{https://github.com/ZhouF-ECNU/ReLKD/blob/main/supplement.pdf}.

\section{Conclusion}
In this paper, we addressed the central challenge of GCD: discovering novel classes within unlabeled data that also contain known categories, while lacking predefined class hierarchy. Existing approaches overlook the inherent relations among classes, limiting their capacity to generalize to novel classes. To overcome this limitation, we proposed ReLKD, an end-to-end framework for GCD that introduces inter-class relation learning and knowledge distillation into GCD for the first time. ReLKD leverages a target-grained module to learn discriminative representations at the granular class level, a coarse-grained module to model implicit class hierarchies, and a distillation module to propagate relational knowledge across class levels, thereby enhancing both representation quality and novel class discovery. Through extensive experiments on four diverse datasets, we demonstrated that ReLKD consistently improves novel class discovery and overall classification performance.
%, setting a foundation for future advancements in open-world recognition and category discovery.